%% file: main.tex
\documentclass{article}
\usepackage{ijcai19}

\usepackage{times}
\usepackage{soul}
\usepackage{url}
\usepackage[hidelinks]{hyperref}
\usepackage[utf8]{inputenc}
\usepackage[small]{caption}
\usepackage{graphicx}
\usepackage{amsmath}
\usepackage{booktabs}
\usepackage{amssymb}
\usepackage{booktabs}
\usepackage{subfig}
\usepackage{enumerate}
\usepackage{siunitx}
\usepackage{multirow}
\usepackage{algorithm,algpseudocode}
\usepackage{color}
\newfloat{algorithm}{t}{lop}
\usepackage{hyperref}

\newcommand*\targ[1]{\overline{#1}}

\title{An Actor-Critic-Attention Mechanism for Deep Reinforcement Learning in Multi-view Environments}

\author{
Elaheh Barati$^1$\footnote{Contact Author}\And
Xuewen Chen$^2$\\
\affiliations
$^1$Department of Computer Science, Wayne State University, Detroit, MI, USA\\
$^2$AIWAYS AUTO, Shanghai, China\\
\emails
elaheh.barati@wayne.edu,
xuewen.chen@ai-ways.com
}

\begin{document}

\maketitle

\begin{abstract}
\input{sections/abstract}
\end{abstract}

\input{sections/introduction}

\input{sections/related-work}

\input{sections/method}

\input{sections/experiments}

\input{sections/conclusion}


\bibliographystyle{named}
\bibliography{ref}

\end{document}

%% file: sections/abstract.tex
In reinforcement learning algorithms, leveraging multiple views of the environment can improve the learning of complicated policies. In multi-view environments, due to the fact that the views may frequently suffer from partial observability, their level of importance are often different. In this paper, we propose a deep reinforcement learning method and an attention mechanism in a multi-view environment. Each view can provide various representative information about the environment. Through our attention mechanism, our method generates a single feature representation of environment given its multiple views. It learns a policy to dynamically attend to each view based on its importance in the decision-making process. Through experiments, we show that our method outperforms its state-of-the-art baselines on TORCS racing car simulator and three other complex 3D environments with obstacles. We also provide experimental results to evaluate the performance of our method on noisy conditions and partial observation settings. 

%% file: sections/introduction.tex
\section{Introduction} 
Distributed reinforcement learning algorithms~\cite{mnih2016asynchronous,barth2018distributed} have been proposed to improve the performance of the learning algorithm by passing copies of the environment to multiple workers. In these methods, although different workers receive different copies of the environment, observations provided by all these copies are from the same sensory input. In other words, the adoption of multiple workers in these works is rather to increase the training reward and earlier convergence, not to collectively increase the amount of observable information from the environment through multiple sensory inputs. 

In an environment such as autonomous driving, an agent may have multiple views of the environment through sensory inputs such as cameras, radars, and so on. These input resources may provide different representative information about the velocity, position, and orientation of the agent. In a realistic environment, observability can be typically partial on account of occlusion from the obstacles or noise that affect the sensors in the environment. Particularly, in environments with pixel-only input, using only one camera view can result in failure, since locating a single camera in a position that can capture both the targets as well as details of agents body is difficult~\cite{tassa2018deepmind}. On the other hand, sensory inputs with less importance can sometimes provide observations which are vital for achieving rich behavior. 
Therefore, it is desirable to incorporate multiple sensory inputs in the decision-making process according to their importance and their provided information at each time. 
The incorporation of multiple sensory inputs in a reinforcement learning environment potentially reduces the sensitivity of policies to an individual sensor and makes the system capable of functioning despite one or more sensors malfunctioning.
Since sensors can provide diverse views of the environment, and they are likely to be perturbed by different noise impacts, a policy is required to attend to the views accordingly.

In this paper, we propose an attention-based deep reinforcement learning algorithm that learns a policy to attend to different views of the environment based on their importance. 
Each sensory input, which provides a specific view of the environment, is assigned to a worker. 
We employ an extension of the actor-critic approach~\cite{silver2014deterministic} to train the network of each worker and make the final decision through the integration of feature representations provided by the workers based on the attention mechanism.
This attention mechanism makes use of critic networks of the workers. Since these critic networks provide signals regarding the amount of salient information supplied by their corresponding views, we employ these signals in the attention module to estimate the amount of impact that each of these views should have in the final decision-making process.
We decouple training of the model, so that view
dependent and task-dependent layers of the network are trained under different strategies.

Unlike distributed reinforcement learning algorithms such as~\cite{mnih2016asynchronous,barth2018distributed}, our method employs different views of the environment as well as different exploration strategies to provide diversity. On the other hand, through the attention mechanism, our method can prevent its decision-making process from being perturbed by degraded views of the environment.
Our main contribution is hence three-fold:
(1) To the best of our knowledge, we are the first to propose a deep reinforcement learning algorithm and a stabilized training process that leverages multiple views of an environment. 
(2) We propose a mechanism to attend to multiple views of the environment with respect to their importance in the decision-making process.
(3) We propose an effective solution to reduce the impact of noise and partial observability on sensory inputs.

%% file: sections/related-work.tex
\section{Related Work} 
In reinforcement learning (RL) tasks that need to deal with continuous action space, it is usually required to maintain an explicit representation of the policy. Multiple approaches have successfully learned policies in continuous domains such as Deep Deterministic Policy Gradient (DDPG)~\cite{lillicrap2015continuous}, proximal policy optimization (PPO)~\cite{schulman2017proximal} and their variants D3PG and D4PG~\cite{barth2018distributed}, and DPPO~\cite{heess2017emergence} in distributed settings. 
~\cite{mnih2016asynchronous} proposed asynchronous advantage actor-critic (A3C) algorithm as a variant of policy gradient that can be applied to continuous action spaces. A3C uses multiple parallel workers (agents) that are fed with the same copy of the environment to explore the state spaces.~\cite{mnih2016asynchronous} investigated maximizing diversity by adopting different policies for the exploration of the environment by different workers. 
~\cite{lillicrap2015continuous} introduced DDPG algorithm, which is based on an actor-critic architecture for representing the policy and value function and makes use of replay buffer to stabilize learning.  Recently,~\cite{barth2018distributed}, proposed D3PG and D4PG, as distributed versions of DDPG. Under ApeX framework~\cite{horgan2018distributed}, D3PG decouples acting and learning and shares the replay buffer among parallel actors. At each time step, a worker samples a batch of transitions from the shared replay buffer and computes the gradients. Distributed Distributional DDPG (D4PG)~\cite{barth2018distributed} is similar to D3PG except it uses the categorical distribution to model the critic function.

In environments with multiple agents, an RL model can incorporate interaction between multiple agents in competitive or cooperative settings. By increasing the number of agents, the problem complexity enhances exponentially that makes applying traditional RL approaches in these environments to be infeasible. To address this challenge, deep multi-agent RL approaches have emerged~\cite{nguyen2018deep,palmer2018lenient,barati2019attention}. Multi-agent RL approaches use different strategies such as optimistic and hysteretic Q-function updates~\cite{lauer2000algorithm,barbalios2014robust,omidshafiei2017deep}. These methods are in cooperative settings where the actions of other agents are made to improve collective reward.~\cite{palmer2018lenient} applied leniency to deep Q-network~\cite{mnih2015human} in a cooperative setting with noisy observations. 
However, there are a number of multi-agent RL methods that are trained in competitive or mixed settings~\cite{bansal2017emergent}.

In this work, we take into account different views of an environment provided by different cameras in which each camera is considered as a worker. Similar to multi-agent systems, there is a common environment that provides the observations of the workers, and if one or multiple workers fail, the remaining workers can take over the task. Moreover, workers help each other to improve their performance in decision making. Our method differs from the multi-agent methods such as~\cite{lowe2017multi,peng2017multiagent} in that all workers take the same action as the final decision while in multi-agent systems, each agent can take its own action. Besides, our method uses the worker's reward to determine the importance of that worker in the final decision-making process. 
Similar to distributed methods~\cite{barth2018distributed,mnih2016asynchronous}, our method achieves a better training performance of its deep network by utilizing multiple workers.
However, in our method, each worker has a distinct view of the environment, while in the distributed methods all workers are fed with the same view of the environment. 

%% file: sections/method.tex
\section{Method}

\input{sections/method-subsections/intro}
\input{sections/method-subsections/crl}

%% file: sections/method-subsections/intro.tex
We propose a method to train policy networks by leveraging multiple observations of the same environment.
We adopt multiple sensory inputs that each provides a distinct observation of the RL environment. In the first step of our learning process, we train multiple workers that each selects an action given a single view of the environment, while in the second step, a single agent learns a policy given all the views of the environment in the global network to make the final decision. 

%% file: sections/method-subsections/crl.tex
\subsection{Attention-based RL Framework}
\begin{figure}[!t]
 \centering
 \includegraphics[width=\columnwidth]{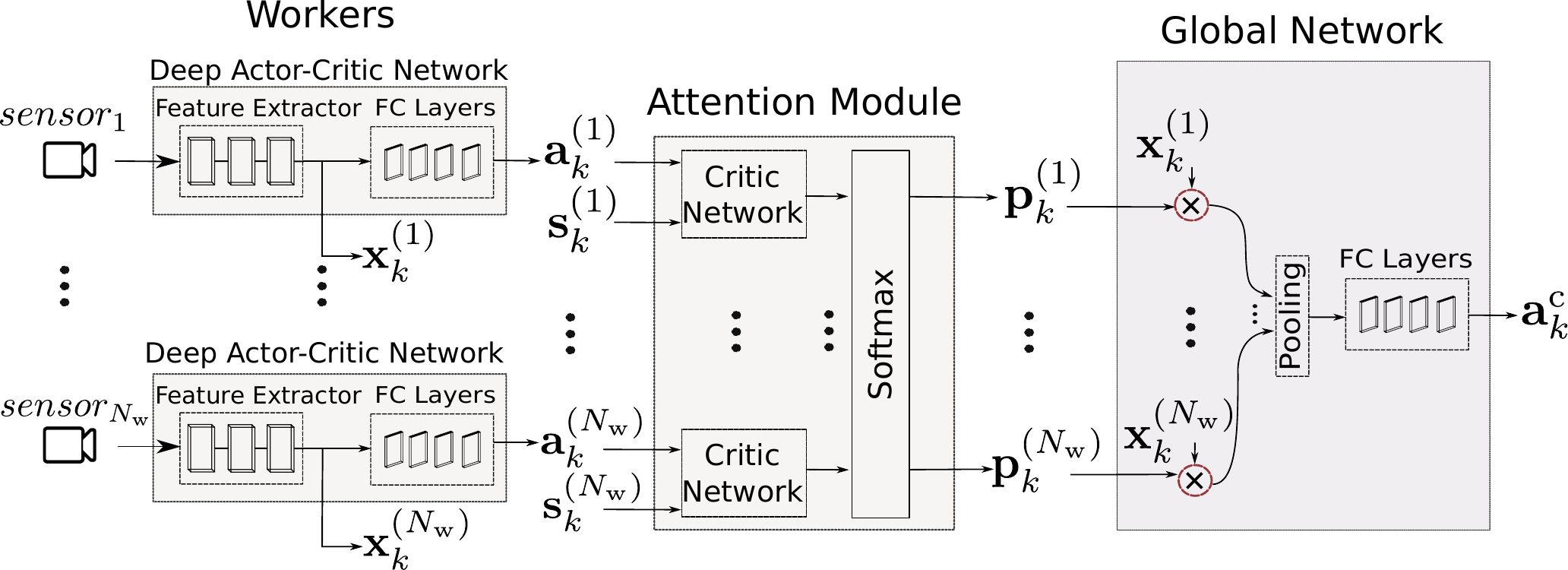}
 \caption{The architecture of the deep network that leverages the attention mechanism in its global network. ${\bf p}^{(w)}_k$ is the weight of worker $w$ obtained from the attention module according to the importance of its view.}\label{fig:collaborative-model}
\end{figure}

Our model (depicted in Figure~\ref{fig:collaborative-model}) is trained in two stages and contains the following two components: (1) the workers' networks that are separately trained at the first stage and retrained at the second stage and (2) the global network that is trained at the second stage by using the feature representations obtained from the workers' networks. 
The decisions made by the workers and their corresponding states are leveraged in an attention module to measure the importance of their associated views. The workers' networks, the global network, and the attention mechanism rely on an actor-critic architecture. 

By utilizing an attention weighted representation as introduced in~\cite{bahdanau2014neural}, our method incorporates the importance of the views in computing a unit representation of the environment (${\bf x}_k$). The attention module has inputs from the critic networks of the workers and gives the highest weight to a view that its corresponding critic network computes the highest Q-value. We use a softmax gate function to compute the attention weights (${\bf p}^{(w)}_k$) as:
\begin{equation}\label{eq:pool}
 {\bf x}_k = \sum_{w=1}^{N_{\rm w}}{{\bf p}^{(w)}_k \odot {\bf x}^{(w)}_k} = \sum_{w=1}^{N_{\rm w}}{\frac{\exp(g_w f_w)}{\sum_{l}{\exp(g_lf_l)}} \odot {\bf x}^{(w)}_k}\; ,
\end{equation}
where $\odot$ stands for Hadamard product, ${\bf x}^{(w)}_k$ is the feature representation obtained by worker $w$, and $g_w$ is a parameter of the model.
In the above equation, $f_w$ is obtained from the output of critic network learned for each worker $w$ separately, i.e.,
\begin{equation}\label{eq:critic}
 f_w = Q(s^{(w)}, a^{(w)})\; .
\end{equation}
At each time step $k$, the attention mechanism generates a positive weight ${\bf p}^{(w)}_k$ for each worker which can be interpreted as the probability that the worker $w$ makes a correct decision. The attention weight ${\bf p}^{(w)}_k$ determines the relative importance of worker $w$ in blending the feature vectors $\{{\bf x}^{(w)}_k| w=1,2,\ldots,N_{\rm w}\}$ together. 

We aim to promote the behavior of each worker by comparing its selected action with the actions selected by all the other workers.
To do so, we modify the reward at the second stage of training by introducing a penalty term that depends on the actions selected by all the workers ($A_k$) as:
\begin{align}\label{eq:r_c}
 r^{\rm c}_k &= r_k - \gamma_{\rm r}{\delta(A_k)}\nonumber\\
 &=r_k - \gamma_{\rm r}\frac{1}{N_{\rm w}} \sum_{w=1}^{N_{\rm w}} \delta^{(w)}(A_k)
\end{align}
where $r_k$ is the original reward, $\gamma_{\rm r}$ is a constant that provides a trade-off between the deviation of actions and the original reward, and $A_k$ is the action matrix with columns ${\bf a}^{(w)}_k$. In~\eqref{eq:r_c}, $\delta(A_k)$
is a deviation function that depends on the variation of action ${\bf a}^{(w)}_k$ of worker $w$ from the average of actions selected by the other workers in the network ($\delta^{(w)}(A_k)$), i.e.,
\begin{align}\label{eq:delta0}
\delta^{(w)}(A_k) = \bigg\vert\bigg\vert {\bf a}^{(w)}_k - \frac{1}{N_{\rm w}-1}\sum_{\substack{v=1\\v\neq w}}^{N_{\rm w}} {\bf a}^{(v)}_k\bigg\vert\bigg\vert^2 \;.
\end{align}
In~\eqref{eq:delta0}, the first term is the action of worker $w$ while the second term is the average action of other workers.
This assumption is valid if the majority of views provide enough information to make the right decision. Some of the views can have lower quality and if we train them individually their corresponding workers might not be trained properly. 
In the case that all workers have trained sufficiently, the deviation $\delta(A_k)$ becomes close to zero, while by experiencing weak training performance from a worker, we get a higher $\delta(A_k)$ and a lower reward $r^{\rm c}_k$. Therefore, the penalty term in~\eqref{eq:r_c} enforces improvement in the training of the workers that are yet to be trained sufficiently. We provide more details regarding the training of the workers in Section~\ref{sec:setup}. We consider $\gamma_{\rm r}$
in~\eqref{eq:r_c} to be $0.1$ in our experiments.
While our method utilizes the actions of the workers in computing the modified reward, these actions do not determine the final action. As described earlier, the final action is determined by the global network from an aggregation of feature representations obtained from the workers. 

Given the state ${\bf s}^{\rm c}_k$ that we set to be ${\bf x}_k$ from~\eqref{eq:pool} and the modified
reward $r^{\rm c}_k$ from~\eqref{eq:r_c}, the parameters of the network including weights of fully-connected (FC) layers of global network ($\theta^{\rm c}$) and workers ($\{\theta^{(w)}\}_{w=1}^{N_{\rm w}}$) and the weights of attention module are trained by using an actor-critic algorithm~\cite{lillicrap2015continuous}. 
At each time step $k$, we stack the modified reward $r^{\rm c}_k$ on the replay buffer, forming a tuple ${\prec} {\bf s}^{\rm c}_k, {\bf a}^{\rm c}_k, r^{\rm c}_k,
{\bf s}_{k+1}^{\rm c}{\succ}$. Then, we sample a random mini-batch from the replay buffer to train critic network by minimizing the following loss function:
\begin{align}
  {\mathcal{L}}(\theta^{Q^{\rm c}}_k) = {\mathbb{E}}\bigg(
  r^{\rm c}_k +
  \gamma \overline {Q}({\bf s}^{\rm c}_{k+1},\targ\mu^{\rm c}({\bf s}^{\rm c}_{k+1};\theta^{\targ\mu^{\rm c}});\theta^{\targ Q^{\rm c}})- \nonumber
  \\
  Q({\bf s}^{\rm c}_k,{\bf a}^{\rm c}_k;\theta^{Q^{\rm c}}_k) \bigg)^2
 \;.
\end{align}
We update the actor network at each time step with respect to the set of parameters $\theta^{\mu^{\rm c}}_k$ by using sampled policy gradient:
\begin{align}
  \nabla_{\theta^{\mu^{\rm c}_k}}J = {\mathbb{E}}\bigg(\nabla_{\bf a} Q({\bf s},{\bf a};\theta^{Q^{\rm c}}_k)|_{{\bf s}={\bf s}^{\rm c}_k , {\bf a}=\mu^{\rm c}({\bf s}^{\rm c}_k)} \nonumber
  \\
  \nabla_{\theta_\mu^{\rm c}}\mu^{\rm c}({\bf s};\theta^{\mu^{\rm c}}_k)|_{{\bf s}={\bf s}^{\rm c}_k}\bigg)\;.
\end{align}

We construct the exploration policy by adding a Gaussian noise to the selected action through the policy $\pi(\bf x)$ of the workers and the global network as
$
{\bf a} = \pi(\bf x) + {\displaystyle \epsilon}\: \mathcal{N}(0, 1)
$.
To promote the behavior of each worker, we control the value of $\displaystyle \epsilon$ for that worker based on the deviation of its actions from the average of actions selected by the other workers. To do so, we choose the value of $\displaystyle \epsilon^{(w)}$ for
worker $w$ to be linearly proportional to the average of $\delta^{(w)}(A_k)$ in $T_{\rm w}$ instances. We choose $T_{\rm w}=1000$ in the experiments. We also compute $\displaystyle \epsilon^{\rm c}$ for exploration during the second stage of training from the average $\displaystyle \epsilon^{(w)}$ of all workers.

We train the network in $M$ iterations. Each of these iterations contains $M_{\rm w}$ episodes of training workers (first stage of training) and $M_{\rm c}$ episodes of training the global network and retraining the FC layers of workers (second stage of training). $M$, $M_{\rm w}$, and $M_{\rm c}$ are hyper-parameters of the model. Before beginning the training of the global network, the workers are trained separately. $M_{\rm w}$ should be large enough such that the decision made by the majority of the workers can be used in the global network to improve the behavior of faulty workers.
Following this strategy, we train the task-dependent layers of workers (fully-connected layers) by taking all workers into account; However, we train the view-dependent layers (convolutional layers) of a worker independent from the other workers.

%% file: sections/experiments.tex
\section{Experiments}
\subsection{Baselines}
We compare the performance of our method, called Attention-based Deep RL (\texttt{ADRL}), with the performances of state-of-the-art actor-critic based methods that are designed to work on continuous action spaces. \texttt{DDPG}~\cite{lillicrap2015continuous}, \texttt{D3PG}~\cite{barth2018distributed}, 
\texttt{PPO}~\cite{schulman2017proximal}, and \texttt{A3C}~\cite{mnih2016asynchronous} are four of the baselines in this work. For \texttt{D3PG} and \texttt{A3C}, we dedicated the same number of workers as \texttt{ADRL} (4 workers) to obtain a stabilized training of the network. However, in \texttt{D3PG} and \texttt{A3C}, we fed the copies of a single view of an environment into their multiple workers. In other words, each worker interacts with its copy of the environment. However, all the workers use the same camera (front view camera) to obtain observations from
their own copy of the environment. We also propose four extensions to \texttt{DDPG} called \texttt{ACT-AVG}, \texttt{ACT-CNT}, \texttt{ACT-MJV}, and \texttt{FT-COMB} and utilize them to provide ablation studies to verify the design choices in the proposed framework.
These four extensions of \texttt{DDPG} adopt multiple workers, each with a different view of an environment, trained by using \texttt{DDPG}. Inspired from~\cite{wiering2008ensemble}, during the training of \texttt{ACT-AVG}, \texttt{ACT-CNT}, and \texttt{ACT-MJV}, the final action is obtained from a combination of actions that the workers take. In \texttt{ACT-AVG}, we take an average of decisions made by workers as the final decision. In \texttt{ACT-CNT}, we find the action vector with the least Euclidean distance from other action vectors. In \texttt{ACT-MJV}, we divide every dimension of the action space into equally distant ten bins and find the bin that contains the maximum number of actions taken by the workers. 
However, \texttt{FT-COMB} is different from others as its states are obtained from the combined features of all views provided by different workers. The final representation is the concatenation of all views features. 
Unlike \texttt{DDPG}, \texttt{D3PG}, \texttt{PPO}, and \texttt{A3C} which work with only a single view of the environment, \texttt{ACT-AVG}, \texttt{ACT-CNT}, \texttt{ACT-MJV}, \texttt{FT-COMB}, and \texttt{ADRL} make use of multiple views in which each view is represented by high-dimensional raw pixels.

\subsection{Experimental Setup}
\paragraph{TORCS Environment.}
We verified our method for autonomous driving on an open-source platform for car racing called TORCS~\cite{wymann2000torcs}. 
We used six different tracks: five tracks for training named in TORCS version 1.3.4 as:
(1) \textit{CG Speedway number 1},
(2) \textit{CG track 2},
(3) \textit{E-Track 1},
(4) \textit{E-Track 2}, and
(5) \textit{Brondehach},
and one track (\textit{Wheel 2}) for testing. We modified road surface textures in TORCS to have the same lane types for all training and testing tracks. In
our experiments, we consider the following actions: acceleration, braking, and steering.
Our objective is to maximize the velocity of the car on the longitudinal axis of the road,
minimize the velocity of the car on the lateral axis, 
and minimize the distance of the car from the road center.
Similar to~\cite{lillicrap2015continuous}, we define the reward function $r_k$ for an agent that takes the action ${\bf
 a}_k$ at time-step $k$ to move from a state to another.

\begin{figure}[!t]
 \centering
 \renewcommand{\arraystretch}{0.1}
 \begin{tabular}{@{}l@{\hspace{1pt}}l@{\hspace{1pt}}l@{\hspace{1pt}}l@{\hspace{1pt}}l@{}}
  \includegraphics[width=0.19\columnwidth]{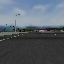} &
  \includegraphics[width=0.19\columnwidth]{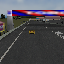}                      &
  \includegraphics[width=0.19\columnwidth]{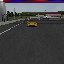}                      &
  \includegraphics[width=0.19\columnwidth]{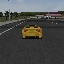} &
  \includegraphics[width=0.19\columnwidth]{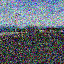}
  \\
 \end{tabular}
 \caption{Four camera views used in this paper plus one with perturbed image
  sample. In TORCS version 1.3.4, these four camera views (from left to right)
  are named:
   (1) \textit{inside car (front view)},
   (2) \textit{behind far},
   (3) \textit{behind near}, and
   (4) \textit{car behind} (distance$=30$).
  The last camera view belongs to Camera 1 but perturbed
  with random Gaussian noise with variance 0.02. }\label{fig:cameras}
\end{figure}

\paragraph{MuJoCo Environment.} 
We also compare the performance of our method with the baselines on three challenging tasks with the continuous action space Ant-Maze, Hopper-Stairs, and Walker-Wall developed in the MuJoCo physics simulator~\cite{todorov2012mujoco}.
In Ant-Maze task, a 3D maze contains five walls, and a MuJoCo ant must navigate from start to the goal state. 
Hopper-Stairs task includes three upward and three downward stairs in a 3D environment where the MuJoCo hopper agent must hop over them.   
In Walker-Wall task, the 3D environment contains three walls that the MuJoCo walker2d must jump or slide over.  
Since we train \texttt{ADRL} and its baselines, given high-dimensional row pixels, it is computationally expensive to achieve the converge to the stable state for all the methods; Therefore, we simplified the texture by modifying the color of the background, walls and body parts of agents.
The action dimension of the Ant-Maze, Hopper-Stairs, and Walker-Wall are, respectively, $8$, $3$, and $6$. The reward is chosen to encourage the agent to reduce the distance from the goal state and increase the forward velocity. In the Ant-Maze environment, the reward also encourages the agent to keep a distance from the walls. In the Hopper-Stairs, the agent is also rewarded based on the torso height.

\begin{figure}[!t]
 \centering
 \renewcommand{\arraystretch}{0.1}
 \begin{tabular}{@{}l@{\hspace{1pt}}l@{\hspace{1pt}}l@{\hspace{2pt}}l@{\hspace{1pt}}l@{\hspace{1pt}}l@{\hspace{1pt}}l@{\hspace{2pt}}l@{\hspace{1pt}}l@{\hspace{1pt}}l@{\hspace{1pt}}l@{\hspace{1pt}}l@{}}
  \includegraphics[width=0.15\columnwidth]{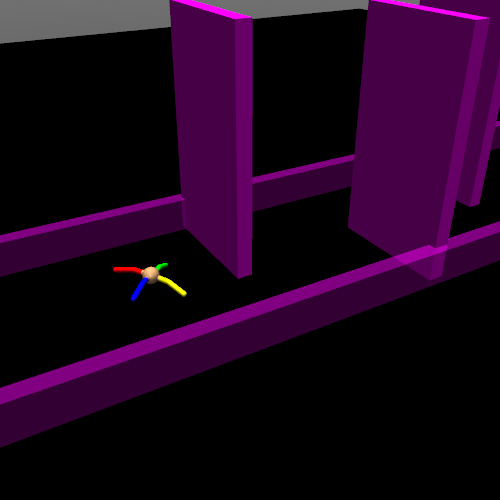} &&
  \includegraphics[width=0.15\columnwidth]{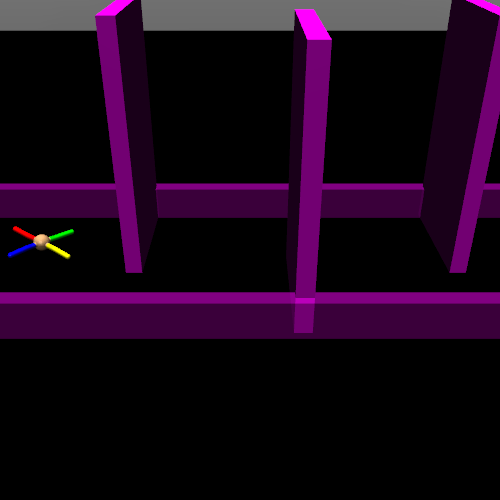} &&
  \includegraphics[width=0.15\columnwidth]{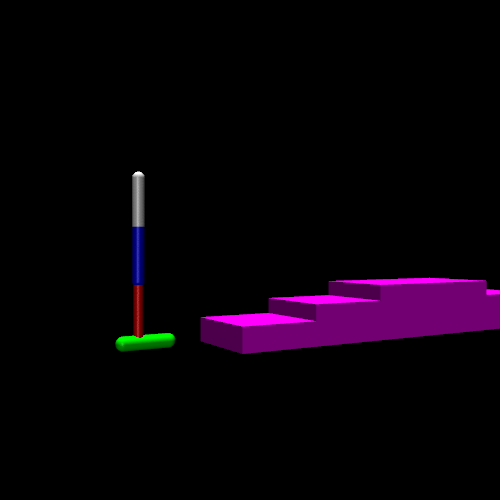} &&
  \includegraphics[width=0.15\columnwidth]{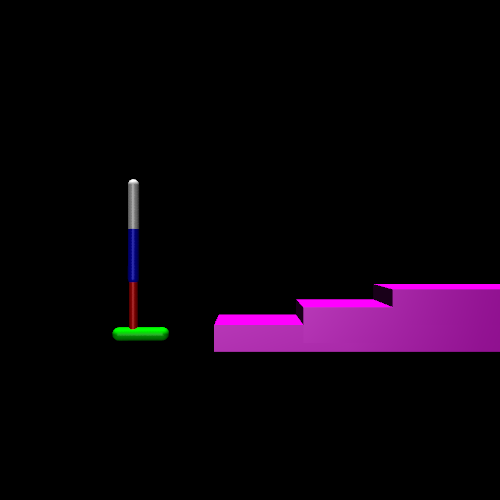} &&
  \includegraphics[width=0.15\columnwidth]{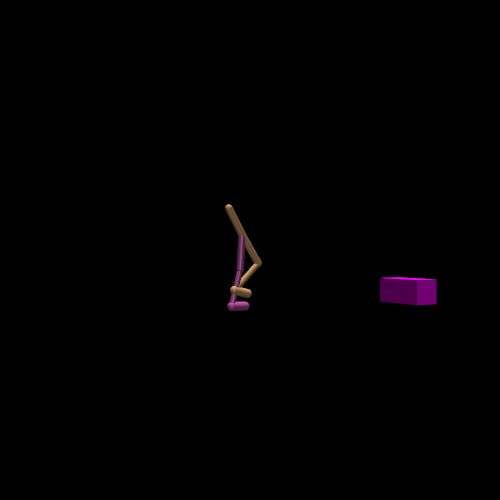} &&
  \includegraphics[width=0.15\columnwidth]{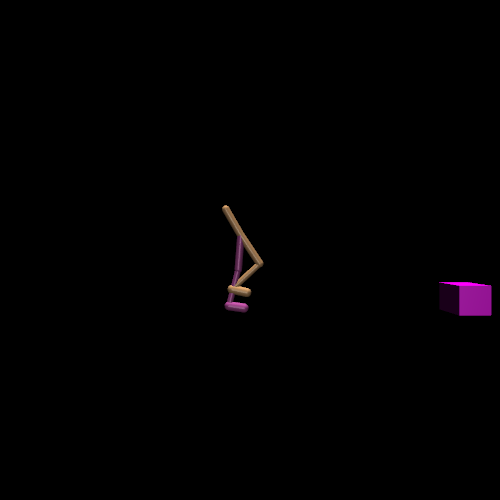}
\\[1pt]
  \includegraphics[width=0.15\columnwidth]{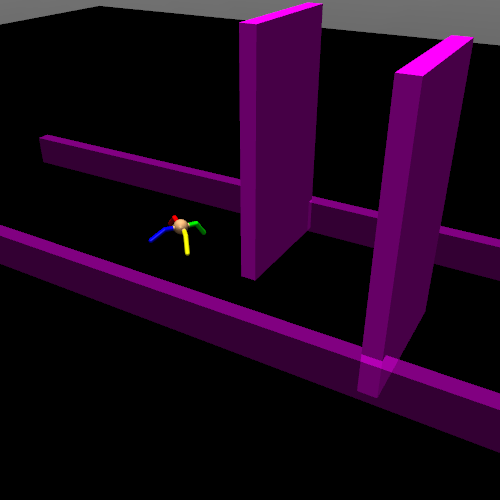} &&
  \includegraphics[width=0.15\columnwidth]{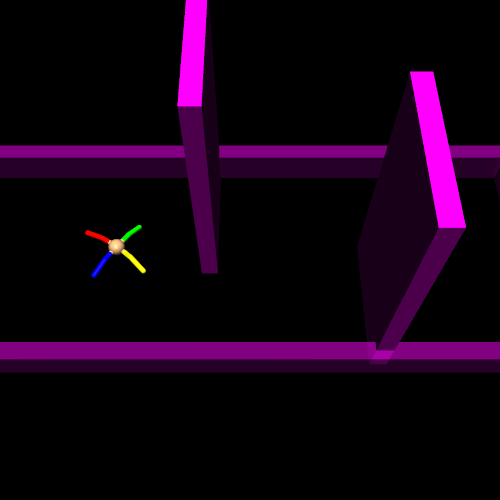} &&
  \includegraphics[width=0.15\columnwidth]{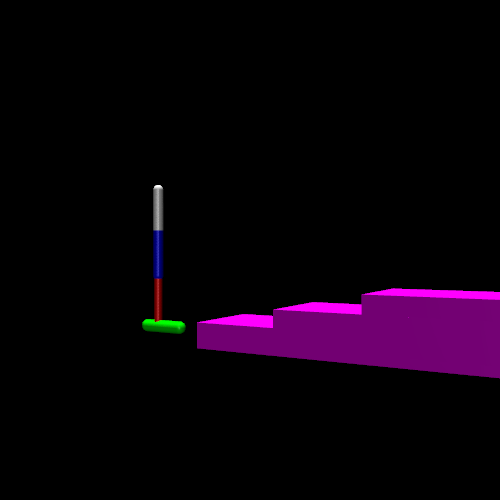} &&
  \includegraphics[width=0.15\columnwidth]{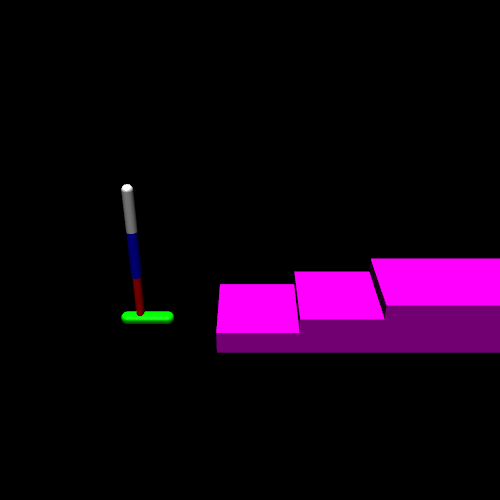} &&
  \includegraphics[width=0.15\columnwidth]{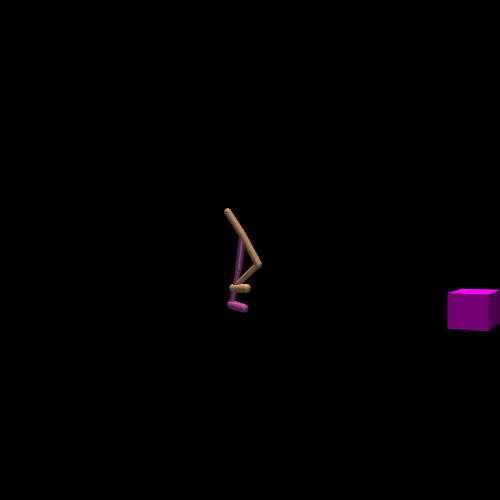} &&
  \includegraphics[width=0.15\columnwidth]{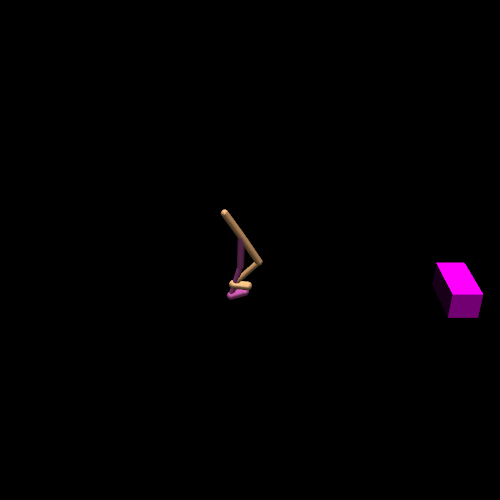}
 \end{tabular}
 \caption{Four camera views: \textit{left view}, \textit{front view}, \textit{right view}, and \textit{top view} (shown in clockwise order) obtained from three MuJoCo based environments: Ant-Maze, Hopper-Stairs, and Walker-Wall. We obtain the left, right and top views of the environment by \ang{30} changes in the camera angle.}\label{fig:mujoco-cameras}
\end{figure}

\begin{figure*}[!t]
 \centering
 \begin{tabular}{l}
  \subfloat[TORCS]
  {\includegraphics[width=.51\columnwidth]{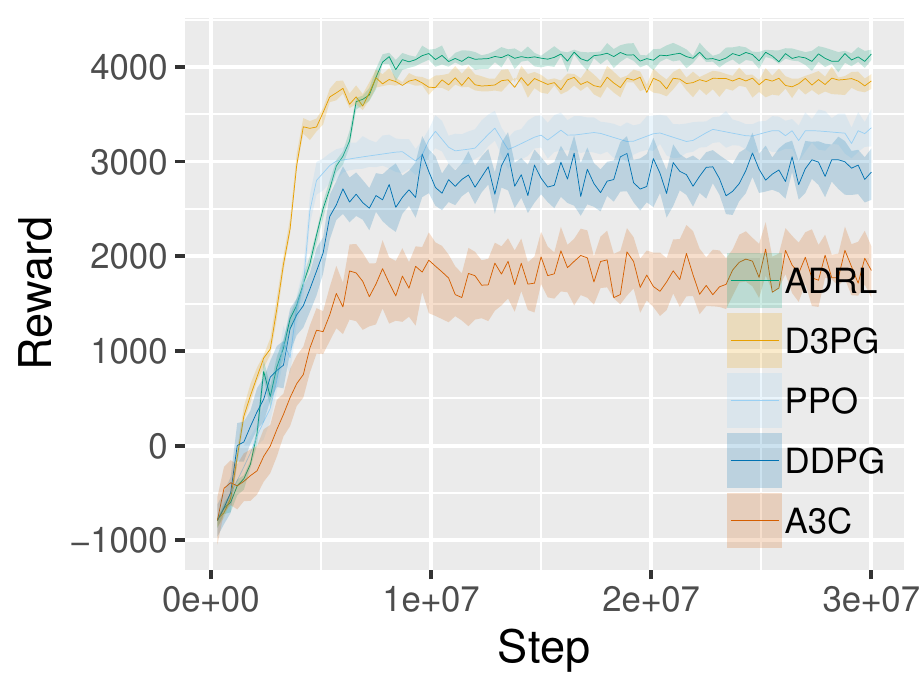}}
  \subfloat[Ant-Maze]{\includegraphics[width=.51\columnwidth]{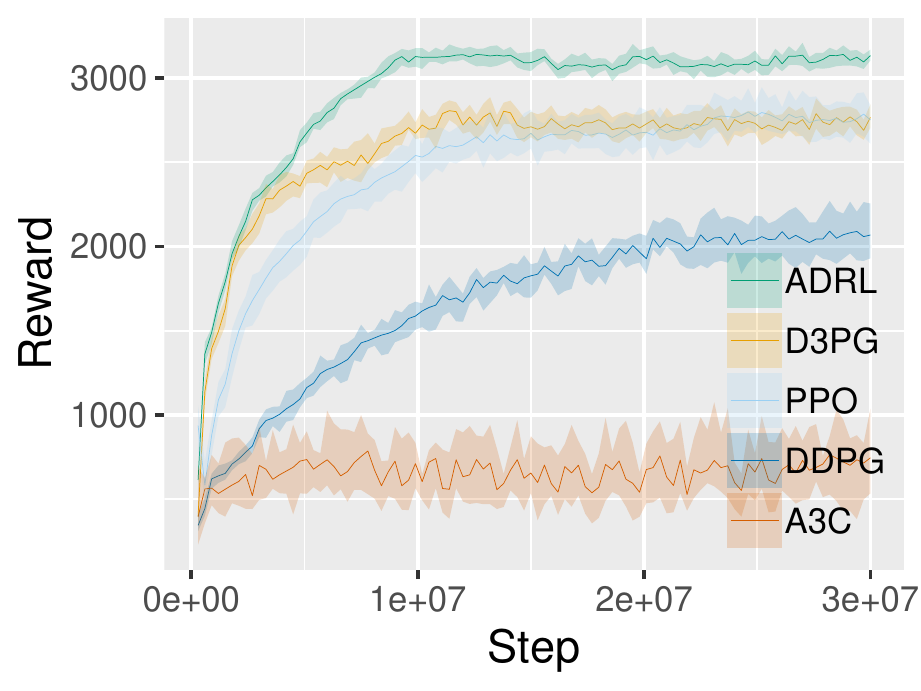}}
  \subfloat[Hopper-Stairs]
  {\includegraphics[width=.51 \columnwidth]{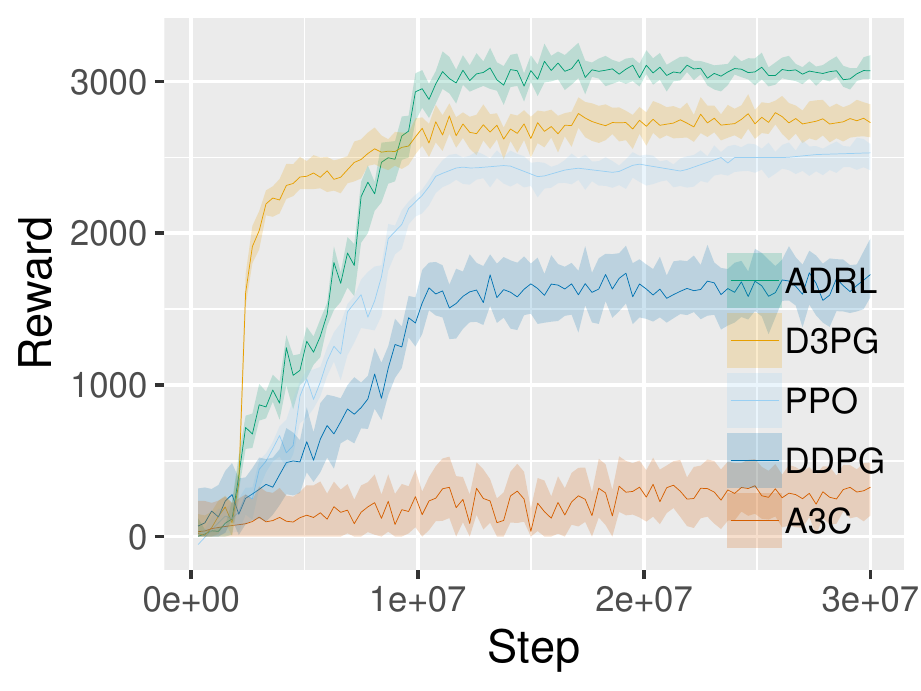}}
  \subfloat[Walker-Wall]{\includegraphics[width=.51\columnwidth]{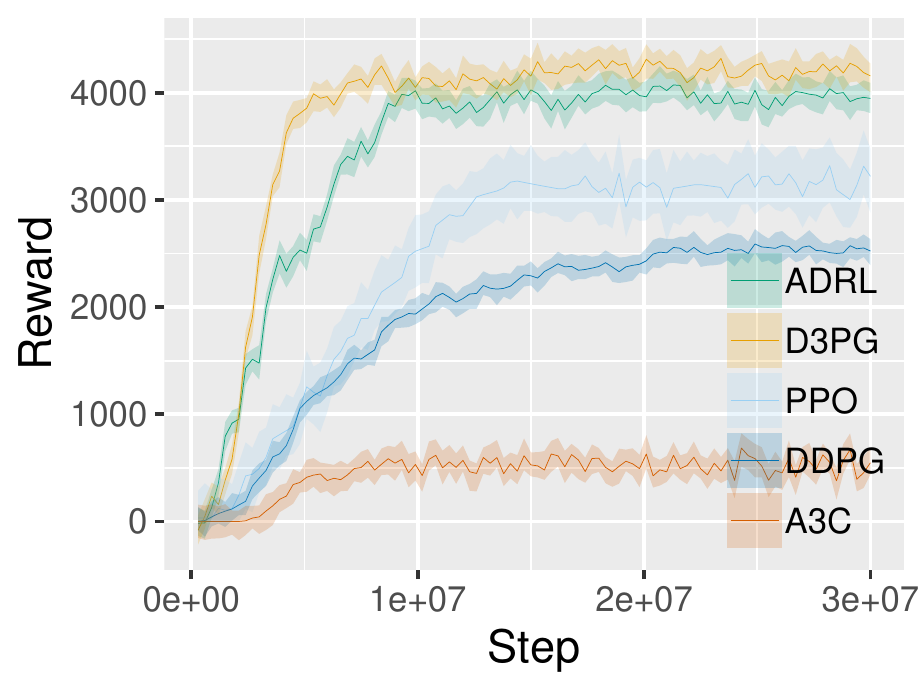}} 
 \end{tabular}
 \caption{Average reward vs.\ training
  step for the methods \texttt{DDPG}, \texttt{D3PG}, \texttt{PPO}, 
  \texttt{A3C}, and \texttt{ADRL}. We obtain the rewards in these figures by averaging rewards obtained from 5 runs. }\label{fig:training}
\end{figure*}

\subsubsection{Multiple Views of Environment}\label{multi-view-env} 
To provide multiple views of environments described by high-dimensional raw pixels, we obtained four different camera views from each environment (depicted in figures~\ref{fig:cameras} and~\ref{fig:mujoco-cameras}).
Particularly, in MuJoCo environment, we obtained these camera views through \ang{30} changes in the camera angle of the front view, while in TORCS environment, we use the pre-defined camera views shown in Figure~\ref{fig:cameras}. In the TORCS environment, the front view camera, unlike the other cameras, does not capture the car itself. 
The Ant-Maze task provides the most diverse camera views because of the complexity of the environment, while the Walker-Wall task provides the least diverse camera views. 

In all the experiments, for the baselines that take a single view of the environment, i.e., \texttt{DDPG}, \texttt{D3PG}, \texttt{PPO}, and \texttt{A3C}, we adopted the \textit{front view} camera. Because we observed that this view provides more useful information that the other views.
We used the four camera views in figures~\ref{fig:cameras} and~\ref{fig:mujoco-cameras} for \texttt{ADRL} and the baselines that take multiple views of the environment, i.e., \texttt{ACT-AVG}, \texttt{ACT-CNT}, \texttt{ACT-MJV}, and \texttt{FT-COMB}. 
We also performed experiments in the cases of (1) having a random Gaussian noise $\mathcal{N}(0,\sigma^2_{E,n})$ on all of the camera views, (2) having one of the cameras perturbed by the noise $\mathcal{N}(0,\sigma^2_{P,n})$ with variance $\sigma^2_{P,n}$ higher than  $\sigma^2_{E,n}$ (see the last view in Figure~\ref{fig:cameras}), and (3) having one or more irrelevant views.

\paragraph{Setup.}\label{sec:setup}We down-sampled the RGB frames into $84\times84$ pixels and normalized them to the range of $[0,1]$. We used three convolutional layers, each with 32 filters as the feature extractor module shown in Figure~\ref{fig:collaborative-model} followed by two fully connected layers with 600 and 400 units. 
All hidden layers are followed by ReLUs. We employed the same architecture for the fully-connected layers in the policy networks of workers and the global network. 
We set the size of replay memory to be $10^5$, the value of discount factor $\gamma = 0.99$, and the learning rates of actor and critic networks to be $10^{-4}$ and $10^{-3}$. To implement and train
the model, we used Tensorflow~\cite{abadi2016tensorflow} distributed machine learning system and applied Adam optimization~\cite{kingma2014adam} for
learning the neural network parameters with the batch size of 32.
For \texttt{ADRL}, we performed $M=100$ iterations of training workers and the global network. 
At the initial training iteration, we dedicated more episodes to train the workers than the global network (i.e., $M_{\rm w}=10\times M_{\rm c}$), and gradually modified the portion of episodes in favor of training the global network such that at the last training iteration $M_{\rm w}=0.1\times M_{\rm c}$.

\subsection{Results and Discussion}
\paragraph{Training Speed Comparison.}
We present the training speed of \texttt{ADRL} and four of its baselines in terms of average reward per step in Figure~\ref{fig:training}. 
At the first stage of training \texttt{ADRL}, multiple workers are trained separately given multiple views of the environment, and the reward of \texttt{ADRL} shown in Figure~\ref{fig:training} is the average reward of all these workers. 
Since \texttt{D3PG} uses all its workers to train the network given a single view of the environment, at the initial steps of training, the reward of \texttt{ADRL} is less than \texttt{D3PG} for all the tasks other than Ant-Maze. However, after convergence, the reward of \texttt{ADRL} is either higher or comparable to \texttt{D3PG} in all the tasks, which is an indication of the advantage of attending to multiple various views of the environment. 
In Ant-Maze task, different camera views (shown in Figure~\ref{fig:mujoco-cameras}) provide significantly diverse information about the environment, and leveraging these views via the attention mechanism in \texttt{ADRL} leads to a significantly higher reward for \texttt{ADRL} in comparison to its baselines.
In Walker-Wall task, the reward of \texttt{ADRL} is slightly less than \texttt{D3PG}, which is due to having a low diversity between the views.

\begin{table*}[!t]
\centering
 \small
 \scalebox{0.78}{
 \subfloat[][TORCS]{
\begin{tabular}{@{}lll@{}}
\toprule
\textbf{Method}  & \textbf{Time (sec)}                      & \textbf{Distance (m)}                    \\ \midrule
                 & \multicolumn{1}{c}{$\mu \; / \; \sigma$} & \multicolumn{1}{c}{$\mu \; / \; \sigma$} \\ \midrule
\textbf{DDPG}    & $0.71\;/\;0.09$                          & $0.64\;/\;0.10$                          \\
\textbf{ACT-AVG} & $0.82\;/\;0.06$                          & $0.68\;/\;0.05$                          \\
\textbf{ACT-CNT} & $0.80\;/\;0.07$                          & $0.68\;/\;0.07$                          \\
\textbf{ACT-MJV} & $0.74\;/\;0.07$                          & $0.66\;/\;0.08$                          \\
\textbf{FT-COMB} & $0.73\;/\;0.12$                          & $0.65\;/\;0.13$                          \\
\textbf{A3C}     & $0.45\;/\;0.11$                          & $0.48\;/\;0.14$                          \\
\textbf{PPO}     & $0.83\;/\;0.04$                          & $0.71\;/\;0.05$                          \\
\textbf{D3PG}    & $0.89\;/\;0.03$                          & $0.82\;/\;0.03$                          \\
\textbf{ADRL}    & $0.92\;/\;0.03$                          & $0.90\;/\;0.03$                          \\ \bottomrule
\end{tabular}
 }
 \subfloat[][Ant-Maze]{
   \begin{tabular}{@{}lll@{}}
\toprule
\textbf{Method}  & \textbf{Time (sec)}                      & \textbf{Distance (m)}                    \\ \midrule
                 & \multicolumn{1}{c}{$\mu \; / \; \sigma$} & \multicolumn{1}{c}{$\mu \; / \; \sigma$} \\ \midrule
\textbf{DDPG}    & $0.41\;/\;0.12$                          & $0.43\;/\;0.14$                          \\
\textbf{ACT-AVG} & $0.63\;/\;0.10$                          & $0.57\;/\;0.11$                          \\
\textbf{ACT-CNT} & $0.58\;/\;0.11$                          & $0.55\;/\;0.11$                          \\
\textbf{ACT-MJV} & $0.57\;/\;0.11$                          & $0.53\;/\;0.09$                          \\
\textbf{FT-COMB} & $0.47\;/\;0.17$                          & $0.46\;/\;0.16$                          \\
\textbf{A3C}     & $0.18\;/\;0.12$                          & $0.19\;/\;0.10$                          \\
\textbf{PPO}     & $0.59\;/\;0.06$                          & $0.60\;/\;0.08$                          \\
\textbf{D3PG}    & $0.61\;/\;0.07$                          & $0.58\;/\;0.08$                          \\
\textbf{ADRL}    & $0.76\;/\;0.06$                          & $0.70\;/\;0.05$                          \\ \bottomrule
\end{tabular}
 }
 \subfloat[][Hopper-Stairs]{
 \begin{tabular}{@{}lll@{}}
\toprule
\textbf{Method}  & \textbf{Time (sec)}                      & \textbf{Distance (m)}                    \\ \midrule
                 & \multicolumn{1}{c}{$\mu \; / \; \sigma$} & \multicolumn{1}{c}{$\mu \; / \; \sigma$} \\ \midrule
\textbf{DDPG}    & $0.54\;/\;0.08$                          & $0.52\;/\;0.11$                          \\
\textbf{ACT-AVG} & $0.64\;/\;0.07$                          & $0.60\;/\;0.07$                          \\
\textbf{ACT-CNT} & $0.60\;/\;0.07$                          & $0.57\;/\;0.06$                          \\
\textbf{ACT-MJV} & $0.61\;/\;0.08$                          & $0.58\;/\;0.08$                          \\
\textbf{FT-COMB} & $0.56\;/\;0.18$                          & $0.56\;/\;0.13$                          \\
\textbf{A3C}     & $0.21\;/\;0.12$                          & $0.18\;/\;0.09$                          \\
\textbf{PPO}     & $0.59\;/\;0.04$                          & $0.60\;/\;0.05$                          \\
\textbf{D3PG}    & $0.73\;/\;0.09$                          & $0.76\;/\;0.07$                          \\
\textbf{ADRL}    & $0.82\;/\;0.04$                          & $0.80\;/\;0.03$                          \\ \bottomrule
\end{tabular}
 }
 \subfloat [][Walker-Wall]{
 \begin{tabular}{@{}lll@{}}
\toprule
\textbf{Method}  & \textbf{Time (sec)}                      & \textbf{Distance (m)}                    \\ \midrule
                 & \multicolumn{1}{c}{$\mu \; / \; \sigma$} & \multicolumn{1}{c}{$\mu \; / \; \sigma$} \\ \midrule
\textbf{DDPG}    & $0.40\;/\;0.14$                          & $0.40\;/\;0.15$                          \\
\textbf{ACT-AVG} & $0.48\;/\;0.12$                          & $0.42\;/\;0.14$                          \\
\textbf{ACT-CNT} & $0.43\;/\;0.13$                          & $0.38\;/\;0.14$                          \\
\textbf{ACT-MJV} & $0.42\;/\;0.12$                          & $0.38\;/\;0.13$                          \\
\textbf{FT-COMB} & $0.44\;/\;0.15$                          & $0.39\;/\;0.15$                          \\
\textbf{A3C}     & $0.15\;/\;0.07$                          & $0.14\;/\;0.06$                          \\
\textbf{PPO}     & $0.49\;/\;0.15$                          & $0.47\;/\;0.16$                          \\
\textbf{D3PG}    & $0.71\;/\;0.10$                          & $0.70\;/\;0.11$                          \\
\textbf{ADRL}    & $0.70\;/\;0.12$                          & $0.68\;/\;0.12$                          \\ \bottomrule
\end{tabular}
}
}
 \caption{Performance comparison of \texttt{ADRL} with its
  baselines on TORCS and three MuJoCo environments. $\mu$ and $\sigma$ are the mean and standard deviation of the time/distance that an agent moves before termination computed over 100 runs. The values of time and distance in this table are normalized with the maximum time and distance the agent travels in all of the runs.}\label{table:baselines}
\end{table*}

\paragraph{Comparison with Baselines.} 
We report the performance of \texttt{ADRL} in comparison to its baselines in Table~\ref{table:baselines}. After learning the parameters of all methods in 30 million training steps, we included background noise $\mathcal{N}(0,\sigma^2_{E,n})$ with $ \sigma^2_{E,n} =
0.01$ to all the camera views and capture the time and distance that the agent moves. In Table~\ref{table:baselines}, we
present the mean $\mu$ and standard deviation $\sigma$ of the time and distance that the agent moves, computed over 100 runs. In this table, the values of time and distance are normalized with the maximum time and distance that the agent moves forward before termination. 
\texttt{ADRL} surpasses all the baselines with respect to mean and standard deviation of time and distance on all tasks except Walker-Wall. In comparison with the other tasks, in Walker-Wall, different views have the least amount of diversity; therefore, the performance of \texttt{ADRL} is comparable to its baselines in this task. 
Since \texttt{FT-COMB} does not utilize an attention mechanism, it has its best performance in the Walker-Wall task where there is a small difference in the importance of different views.
\texttt{ADRL} and the baselines \texttt{ACT-AVG}, \texttt{ACT-CNT}, \texttt{ACT-MJV}, and \texttt{FT-COMB} employ multiple workers fed with different views of environment; however, unlike \texttt{ADRL},
these four baselines can only beat \texttt{DDPG} and \texttt{A3C}. This is mainly an indication of improvement due to the attention mechanism and the policy network used in \texttt{ADRL}. As a result of averaging actions made by multiple workers trained on multiple views, we observe that \texttt{ACT-AVG}, \texttt{ACT-CNT}, and \texttt{ACT-MJV} have lower
variances than \texttt{DDPG}; nevertheless, its variance is still higher than \texttt{ADRL}.
\texttt{ADRL} has the highest improvement over \texttt{D3PG} ($>$20\%) in the Ant-Maze task, which has the highest diversity in its views. This suggests that \texttt{ADRL} has higher improvement when there is higher diversity among views. 

\begin{figure}
 \centering
 \includegraphics[width=0.8\columnwidth]{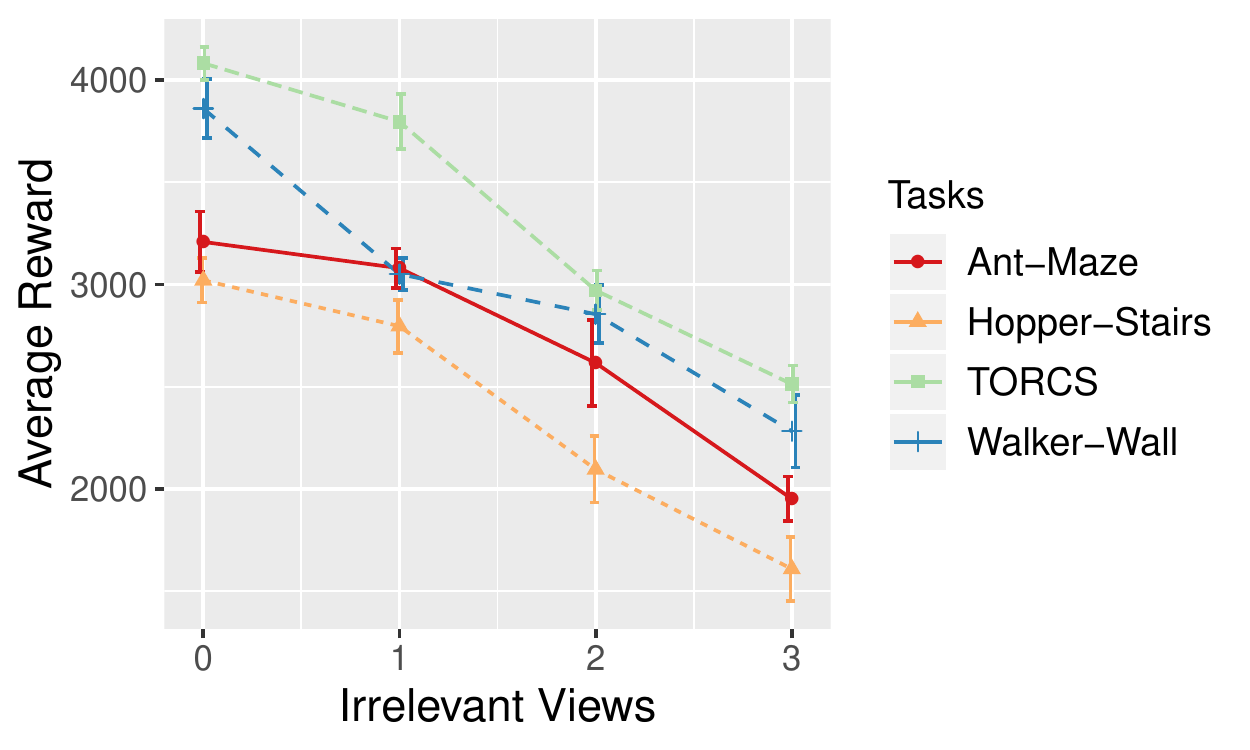}
 \caption{Impact of irrelevant views in average reward.}\label{fig:irrelevant_views}
\end{figure}

\paragraph{Irrelevant views Impact.}
We examine the stability of \texttt{ADRL} in the case of existing a number of irrelevant views. We trained \texttt{ADRL} for the case of having four cameras that are located in the same way described in subsection~\ref{multi-view-env}. After 30 million steps of training, we made camera views irrelevant by randomly changing the camera angles to ones that have no view of the agent. In the TORCS environment, we captured the irrelevant view by a side camera along the road that provides a view of the car only in a very limited period of time. Figure~\ref{fig:irrelevant_views} demonstrates the average testing rewards over 5 runs for the case of having 0, 1, 2, and 3 irrelevant camera views. We observed that for the cases of having at least one relevant camera, the testing rewards were satisfactory. Specifically, by making three out of four cameras irrelevant, the average rewards decrease around 37\% in all the tasks.  This observation indicates that the attention module is able to adjust the weight of the irrelevant camera to reduce deterioration in the average reward. From Figure~\ref{fig:irrelevant_views} 
we observe that by having one irrelevant camera, there is a slight decrease in the average testing reward (around 7\%).
From Figure~\ref{fig:irrelevant_views} and Figure~\ref{fig:training}, we can conclude that \texttt{ADRL} works better than \texttt{DDPG} in the case of having only one relevant camera. In other words, due to the penalty term considered in the training reward of \texttt{ADRL}, each of the workers gets better training than the case of training them individually. 

\paragraph{Noise Perturbation Impact.}
We examine the performance of \texttt{ADRL} in the TORCS environment under a scenario in which one out of four views of the environment comes with less quality due to having a randomly chosen camera perturbed by noise with higher variance than the others. We used
the first four camera views listed in Figure~\ref{fig:cameras} to obtain
the high-dimensional views. In this scenario, we adopted single and multiple views (shown in Table~\ref{table:perturbationscenarios}) to compare
\texttt{ADRL} with the best performing baseline in Table~\ref{table:baselines} that takes multiple different views (i.e., \texttt{ACT-AVG}).
Using only a single view, \texttt{ADRL} does not need any attention mechanism over the views, which makes its performance equivalent to the performance of \texttt{DDPG}.
Table~\ref{table:perturbationscenarios} summarizes the mean and the standard deviation of the time and distance that a car travels until it leaves the road. We added zero-mean Gaussian noise with the variances $0.01$ and $0.05$ to represent high- and low-quality views, respectively. We observe the superior performance of \texttt{ADRL} over \texttt{ACT-AVG} in all the cases in Table~\ref{table:perturbationscenarios}. 
We observe that the worst performance occurs when all the views contain high-variance noise. We can conclude that \texttt{ADRL} is more successful than \texttt{ACT-AVG} in dealing with the case of having a low-quality view.

\begin{table}[!t]
 \centering
\scriptsize
 \setlength{\tabcolsep}{5.5pt}
 \begin{tabular}{@{}llllll@{}}
  \toprule
  \multirow{2}{*}{\tiny \begin{tabular}{@{}c@{}}\textbf{Method/}\\ {\bf Feature Level}\end{tabular}}        & $\sigma^2_{E,n}$ & $\sigma^2_{P,n}$ & $N_{\rm w}$ & \textbf{Time (sec)}                        & \textbf{Distance (m)}                      \\ \cmidrule(l){5-6}
   &                  &                  &             & \multicolumn{1}{c}{$\mu \; / \; \sigma$} & \multicolumn{1}{c}{$\mu \; / \; \sigma$} \\ \midrule
  \textbf{ADRL}        & $0.01$           & ---              & 1           & $0.73\;/\;0.08$                           & $0.67\;/\;0.10$                        \\
  \textbf{ADRL}        & $0.05$           & ---              & 1           & $0.14\;/\;0.04$                            & $0.15\;/\;0.05$                          \\\midrule
  \textbf{ADRL}        & $0.01$           & ---              & 4           & $0.92\;/\;0.03$                           & $0.90\;/\;0.03$                        \\
  \textbf{ACT-AVG}     & $0.01$           & ---            & 4           & $0.82\;/\;0.06$                           & $0.78\;/\;0.05$                        \\
  \textbf{ADRL}        & $0.05$           & ---              & 4           & $0.29\;/\;0.12$                           & $0.32\;/\;0.14$                         \\
  \textbf{ACT-AVG}     & $0.05$           & ---            & 4           & $0.18\;/\;0.07$                           & $0.18\;/\;0.08$                          \\\midrule
  \textbf{ADRL}        & $0.01$           & $0.05$           & 4           & $0.84\;/\;0.08$                           & $0.80\;/\;0.07$                        \\
  \textbf{ACT-AVG}     & $0.01$           & $0.05$           & 4           & $0.77\;/\;0.19$                           & $0.74\;/\;0.15$                        \\
  \bottomrule
 \end{tabular}
  \caption{ Noise perturbation impact on the mean ($\mu$) and standard deviation ($\sigma$) of time and distance a car travels
  without leaving the road by using \texttt{ADRL} and \texttt{ACT-AVG}. We normalize the values of time and distance in this table. $\sigma^2_{E,n}$ and
  $\sigma^2_{P,n}$ are the environment noise and noise on a perturbed
  sensor.}\label{table:perturbationscenarios}
\end{table}

%% file: sections/conclusion.tex
\section{Conclusion}
We proposed an attention-based deep reinforcement learning method for multi-view environments. Our method takes advantage of multiple views of the environment with a different exploration strategy for each view to obtain a stabilized training policy. This method dynamically attends to views of the environment according to their importance in the final decision-making process where we use a critic network designated for each view to measure the importance of the view. Through the experiments, we observed that our method outperforms its baselines that use single
or multiple views of the environment. The experiments also reveal the superior performance of our method in the face of a degraded view of the environment. As future work, we will investigate incorporating memory components in the attention module to model temporal changes in the quality of views of the environment.